\pdfoutput=1

\documentclass[11pt]{article}

\usepackage[final]{acl}

\usepackage{times}
\usepackage{latexsym}

\usepackage[utf8]{inputenc} % allow utf-8 input
\usepackage[T1]{fontenc}    % use 8-bit T1 fonts
\usepackage{hyperref}       % hyperlinks
\usepackage{url}            % simple URL typesetting
\usepackage{booktabs}       % professional-quality tables
\usepackage{amsfonts}       % blackboard math symbols
\usepackage{nicefrac}       % compact symbols for 1/2, etc.
\usepackage{microtype}      % microtypography
\usepackage{xcolor}         % colors
\usepackage{graphicx}
\usepackage{ragged2e}
\usepackage{multirow}
\usepackage{amsmath}
\usepackage{arydshln}
\usepackage{subcaption}
\usepackage{caption}
\usepackage{algorithm}
\usepackage{algpseudocode}
\usepackage{color}
\usepackage{wrapfig}
\usepackage{enumitem}
\usepackage{tikz}
\usepackage{tcolorbox}
\usepackage{verbatim}
\usepackage{fancyvrb}

\title{CityNavAgent: Aerial Vision-and-Language Navigation with \\ Hierarchical Semantic Planning and Global Memory}

\author{
 \textbf{Weichen Zhang},
 \textbf{Chen Gao\textsuperscript{$\ddagger$}},
 \textbf{Shiquan Yu},
 \textbf{Ruiying Peng},
 \textbf{Baining Zhao},
 \\
 \textbf{Qian Zhang},
 \textbf{Jinqiang Cui},
 \textbf{Xinlei Chen\textsuperscript{$\ddagger$}},
 \textbf{Yong Li}
\\
 Tsinghua University, 
\textsuperscript{$\ddagger$}Corresponding Author
\\
\texttt{chgao96@gmail.com}, \texttt{liyong07@tsinghua.edu.cn}, \\
\texttt{chen.xinlei@sz.tsinghua.edu.cn}
}

\begin{document}

\maketitle

\begin{abstract}
Aerial vision-and-language navigation (VLN) — requiring drones to interpret natural language instructions and navigate complex urban environments — emerges as a critical embodied AI challenge that bridges human-robot interaction, 3D spatial reasoning, and real-world deployment.
Although existing ground VLN agents achieved notable results in indoor and outdoor settings, they struggle in aerial VLN due to the absence of predefined navigation graphs and the exponentially expanding action space in long-horizon exploration. In this work, we propose \textbf{CityNavAgent}, a large language model (LLM)-empowered agent that significantly reduces the navigation complexity for urban aerial VLN. 
Specifically, we design a hierarchical semantic planning module (HSPM) that decomposes the long-horizon task into sub-goals with different semantic levels. The agent reaches the target progressively by achieving sub-goals with different capacities of the LLM. Additionally, a global memory module storing historical trajectories into a topological graph is developed to simplify navigation for visited targets.
Extensive benchmark experiments show that our method achieves state-of-the-art performance with significant improvement. Further experiments demonstrate the effectiveness of different modules of CityNavAgent for aerial VLN in continuous city environments. The code is available at \href{https://github.com/VinceOuti/CityNavAgent}{link}.

\end{abstract}

\section{Introduction}
Visual-and-language navigation (VLN) is a fundamental task where an agent is required to navigate to a specified landmark or location following language instructions \cite{anderson2018vision, gu2022vision, gao2024vision}. With the increasing prevalence of unmanned aerial vehicles (UAVs), aerial VLN \cite{liu2023aerialvln} has gained significant attention. This task empowers UAVs to navigate complex, large-scale outdoor environments with language instructions, reducing the cost of human-machine interaction and offering significant advantages in applications like rescue, transportation, and urban inspections.

\begin{figure}[t!]
  \centering
  \includegraphics[width=\linewidth]
  {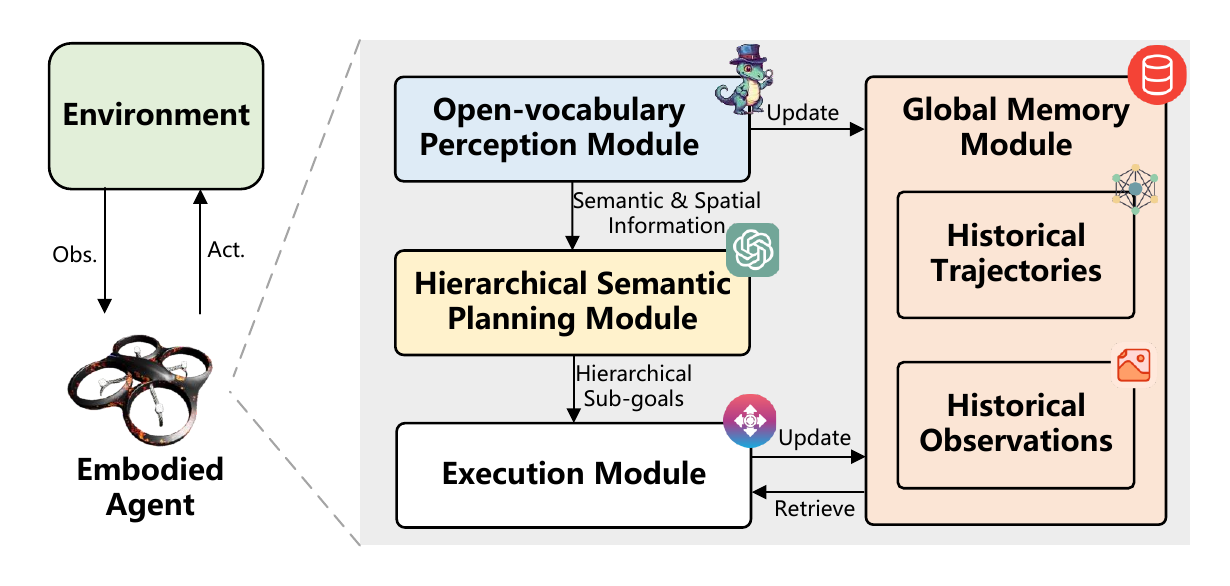}
  \setlength{\abovecaptionskip}{-5pt}
%\vspace{-0.3cm}
  \caption{The overall workflow of CityNavAgent.}
  \vspace{-0.4cm}
  \label{framework}
\end{figure}

Most existing methods primarily address indoor VLN. One approach \cite{anderson2018vision, kurita2020generative, chen2024mapgpt, gao2023adaptive, huo2023geovln, chen2022think} formulates the task in a discrete setting, where agents teleport between nodes in pre-defined topological graphs without motion errors, limiting real-world applicability. Other methods mitigate reliance on pre-defined maps using end-to-end action prediction \cite{krantz2020beyond, raychaudhuri2021language, chen2021topological} or waypoint prediction \cite{hong2022bridging, an2024etpnav, wang2024lookahead}. But the former struggles with scene semantic variations, while the latter fails to adapt to large-scale outdoor scenarios. 
Although some methods \cite{schumann2024velma, liu2024multimodal} extend VLN to outdoor ground navigation, they still rely on pre-defined scene graphs, which are unavailable in aerial settings. STMR \cite{gao2024aerial} introduced a zero-shot LLM-based framework for aerial VLN by constructing an online 2D semantic map, but its failure to incorporate height information leads to high navigation errors.

In this work, we focus on aerial VLN that has a more realistic and challenging setting compared to the previous VLN tasks. In this task, the agent is required to predict the next action or waypoint to approach to the target iteratively in a continuous aerial space. The challenges are two-fold:
\begin{itemize}[leftmargin=*,partopsep=0pt,topsep=0pt]
\setlength{\itemsep}{0pt}
\setlength{\parsep}{0pt}
\setlength{\parskip}{0pt}
    \item \textbf{Complex scene understanding in urban environments} Urban environments exhibit considerably higher object variety than indoor scenes, incorporating extensive infrastructural elements, architectural structures, and natural landscapes. Moreover, the semantic density in urban scenes is highly dynamic. When an agent operates near ground level, the scene exhibits high semantic density, whereas at higher altitudes the semantic becomes markedly sparse. These disparities in object variety and semantic density pose substantial challenges for cross-modal grounding and instruction-related object extraction.
    \item \textbf{Exponential complexity in long-horizon motion planning.} The VLN task can be considered as a Partially Observable Markov Decision Process, where the agent predicts the next action or waypoint based on its current state and the environmental context. However, for the aerial VLN,  the long-horizon navigation requires the agent to predict longer action sequences. Even if the number of available actions per step is limited, the total number of possible action sequences grows exponentially with the planning horizon. Specifically, if there are $m$ actions available at each step, the number of potential action sequences over $n$ steps is approximately $m^n$, which poses a great challenge to the agent’s action planning.
\end{itemize}

In this work, we propose \textbf{CityNavAgent}, which consists of an \textbf{open-vocabulary perception module} and a \textbf{hierarchical semantic planning module (HSPM)} with a \textbf{global memory module} to address the above challenges. 1) To extract the complex and rich semantics in urban environments, the open-vocabulary perception module first utilizes an LLM to caption the scene and extract instruction-related objects through prompt engineering. It then integrates a vision foundation model for open-vocabulary image grounding. 2) To narrow down the possible action space during the motion planning, 
we design HSPM, which decomposes the navigation task into landmark-level, object-level, and motion-level planning, with progressively decreasing semantic abstraction. The planning frequency decreases from low to high levels. The landmark-level planning decomposes the navigation task into a sequence of landmarks to be traversed. The object-level planning reasons about the objects in the scene that lead to these landmarks. The motion-level planning predicts the waypoint and action sequence to reach the semantic target from higher-level planners. Additionally, CityNavAgent incorporates a global memory module to store effective waypoints and trajectories from historical tasks, enhancing long-term navigation performance.

To summarize, the main contributions of this work are as follows:
\begin{itemize}[leftmargin=*,partopsep=0pt,topsep=0pt]
\setlength{\itemsep}{0pt}
\setlength{\parsep}{0pt}
\setlength{\parskip}{0pt}
    \item We focus on the urban aerial VLN task, which is insufficiently explored yet valuable and highly challenging, and introduce \textbf{CityNavAgent}—an LLM-powered agent for zero-shot navigation.
    \item We propose an open-vocabulary perception module that enables the agent to understand the urban scene and HSPM with global memory that reduces the complexity of action planning to address the key challenges.
    \item We conduct extensive experiments on two aerial VLN benchmarks to demonstrate our proposed method in terms of success rate and path following. More ablation studies verify the efficacy of our designed components.
\end{itemize}

\section{Related Works}
\vspace{-0.2cm}
\textbf{Vision-and-language Navigation (VLN)}
VLN is first well defined by R2R \cite{anderson2018vision} which is a navigation benchmark collected in a photorealistic simulator \cite{chang2017matterport3d} with detailed language descriptions and visual observations. Based on R2R, tons of methods \cite{shridhar2020alfred,gao2023adaptive,huo2023geovln,chen2021history,kamath2023new,li2023improving,li2023kerm,chen2022think,guhur2021airbert,qi2021road} are proposed to enable the robots with embodied navigation capacity. Specifically, Kurita et al. \cite{kurita2020generative} proposed a novel generative approach that predicts the instruction distribution conditioned on the action set.
However, R2R and its derivation \cite{ku2020room,jain2019stay} are defined in limited indoor scenes and discrete action spaces which the agent moves within pre-defined topological graphs. This setting yields its practical application in the continuous real-world space.

Krantz et al. \cite{krantz2020beyond} introduced R2R-CE tasks by adapting R2R trajectories for continuous environments. End-to-end methods such as LSTM-based methods \cite{krantz2020beyond,raychaudhuri2021language,liu2023aerialvln}, transformer-based methods  \cite{irshad2022semantically, chen2021topological, krantz2021waypoint}, and reinforcement learning-based methods \cite{wang2018look,wang2020soft} have been explored to improve navigation policies. More recently, waypoint-based methods, such as \cite{hong2022bridging, an2024etpnav,wang2023dreamwalker,wang2024lookahead} have emerged by constructing online maps and waypoints to narrow down the agent's possible locations during navigation. Despite these advancements, end-to-end and waypoint prediction-based methods still face obstacles in adapting to open outdoor environments, primarily due to the differences in spatial structures and the semantic distribution of objects.

In this work, we focus on outdoor aerial VLN \cite{liu2023aerialvln}, a more challenging task with longer navigation paths, more diverse scene semantics, and more complex action spaces. More specifically, unlike ground-level outdoor VLN \cite{chen2019touchdown,schumann2020generating,schumann2024velma} that operate within discrete action spaces, aerial VLN requires agents to navigate through continuous 3D spatial coordinates, which is a more realistic setting for real-world navigation.

\noindent \textbf{LLM for Embodied Navigation}
With the rise of LLMs \cite{touvron2023llama, chiang2023vicuna, achiam2023gpt, brown2020language}, many methods \cite{dorbala2022clip, chen20232, schumann2024velma, shah2023lm} have leveraged their reasoning capabilities for zero-shot VLN. The main challenge for zero-shot LLM-based methods lies in constructing condensed and structured semantic maps of the environment, such as topological graphs, so that LLMs can reason over the semantic information and predict the next waypoint on these maps. Existing works either use pre-defined semantic maps \cite{achiam2023gpt, zhou2024navgpt, chen2024mapgpt} provided by simulators or predict semantic maps within indoor scenes \cite{wang2023dreamwalker}. However, indoor semantic map prediction methods face challenges related to scale and semantic shifts when applied to outdoor environments. STMR \cite{gao2024aerial} proposed an outdoor online 2D semantic map construction pipeline and achieved promising results on aerial VLN. But it fails to leverage the height information of the scene, which is also critical for navigation. In this work, we propose CityNavAgent, which comprises a hierarchical semantic planner that predicts waypoints in outdoor environments in a zero-shot manner, along with a global memory module that stores historical waypoints to enhance long-term navigation.

\section{Problem Formulation}
\label{prob_def}
Given a language instruction $\mathcal{I}$ and the agent's egocentric observation $\mathcal{O}$, the aerial VLN agent has to determine a sequence of action to reach the target location $p_d$ in a continuous 3D space. 
At each action step $t$, the agent follows a policy $\pi$ taking current observation $o_t$ and instruction $\mathcal{I}$ as input to predict the next action $a_t$ and move to location $p_t$ by its kinematic model $\mathcal{F}$, which is given by:
\begin{equation}
\setlength\abovedisplayskip{3pt}%shrink space
\setlength\belowdisplayskip{3pt}
\label{step_policy}
p_t = \mathcal{F}(p_{t-1}, \pi(o_t, \mathcal{I})),
\end{equation}
Given a sequence of action, the agent reaches a final position, and the success probability $P_s$ of reaching the target $p_{d}$ is
\begin{equation}
\setlength\abovedisplayskip{3pt}%shrink space
\setlength\belowdisplayskip{3pt}
\label{step_pos}
% P_{s} = P(||p_n - p_d|| < \epsilon),
P_{s} =P(||\mathcal{F} (\pi(p_0, \mathcal{O}, \mathcal{I}))-p_d|| < \epsilon),
\end{equation}
where $||\cdot||$ is the Euclidean distance and $\epsilon$ is the threshold that indicates if the target is reached. Thus, the goal of VLN is to find a policy $\pi^*$ that maximizes the success rate, given by:
\begin{equation}
\setlength\abovedisplayskip{3pt}%shrink space
\setlength\belowdisplayskip{3pt}
\label{step_pos}
\pi^* = \text{argmax}_{\pi} P_s.
\end{equation}

\section{CityNavAgent}
\label{method}
In this section, we present the workflow of the proposed CityNavAgent for zero-shot aerial VLN in urban environments. As shown in Figure~\ref{framework}, CityNavAgent framework comprises three key modules. 1) The open-vocabulary perception module extracts structured semantic and spatial information from its panoramic observation via a foundation model. 2) The hierarchical semantic planning module (HSPM) leverages LLM to decompose the navigation task into hierarchical planning sub-tasks to reduce the planning complexity and predict the intermediate waypoint for the execution module. 3) The global memory module, represented as a topological graph, stores valid waypoints extracted from historical trajectories to further reduce the action space of motion planning and enhance long-range navigation stability.

\subsection{Open-vocabulary Perception Module}
\label{scene perception module}
%To perceive the open urban environment, the aerial agent has to recognize the objects in the scene and extract their semantic and spatial information. 
To accurately understand complex semantics in the urban environment, we leverage the powerful open-vocabulary captioning and grounding model to extract scene semantic features. Besides, integrating the scene semantics and depth information, we construct a 3D semantic map for further planning.

\begin{figure*}[t!]
  \centering
  \includegraphics[width=0.9\linewidth]{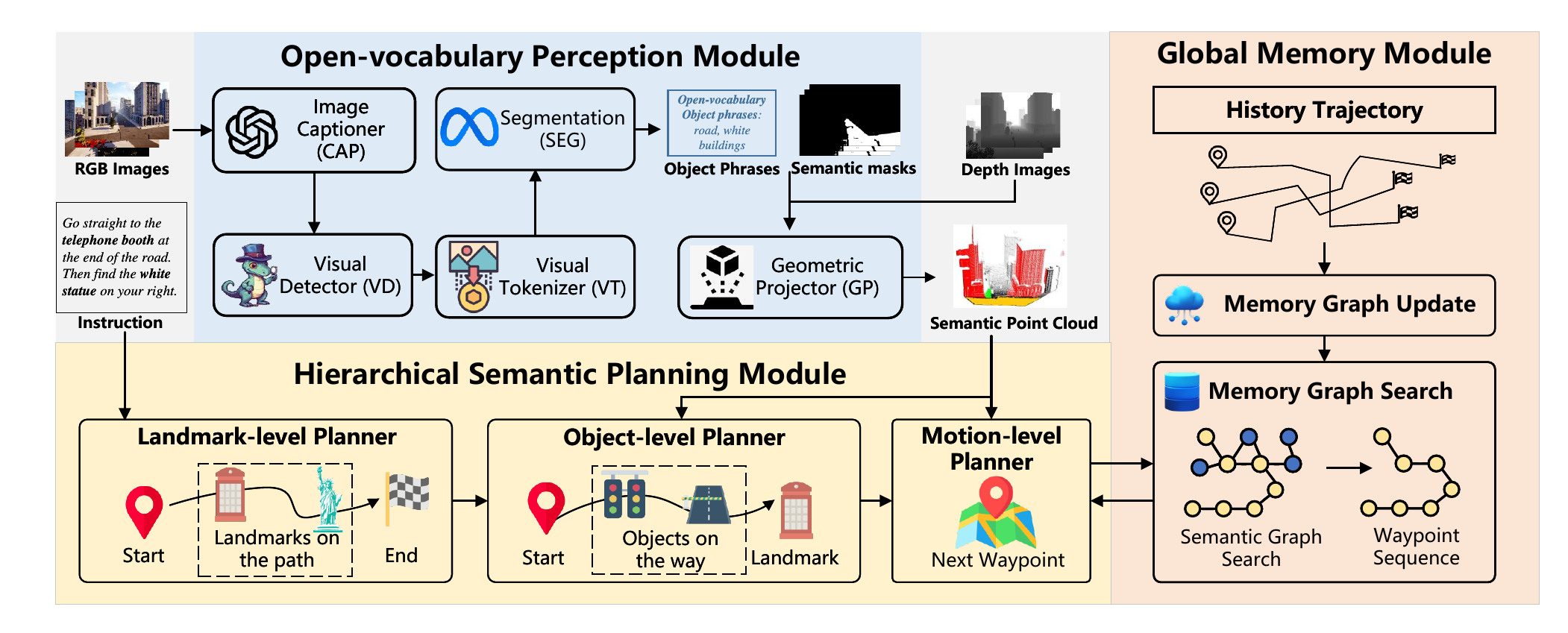}
 %   \vspace{-0.3cm}
 \setlength{\abovecaptionskip}{-3pt}
  \caption{{CityNavAgent consists of three key modules. The open-vocabulary module extracts open-vocabulary objects in the scenes and builds a semantic point cloud of the surroundings. The hierarchical semantic planning module decomposes the original instruction into sub-goals with different semantic levels and predicts the agent's next action to achieve the high-level sub-goals. The global memory module stores historical trajectories to assist motion planning toward visited targets.}}
  \vspace{-0.4cm}
  \label{perception}
\end{figure*}

\noindent \textbf{Scene Semantic Perception}
Extracting urban scene semantics requires robust open-vocabulary object recognition. As shown in Figure~\ref{perception}, given a set of panoramic images $\{I^t_{R_i}\}$ at step $t$, we first use an open-vocabulary image captioner $\text{CAP}(\cdot)$ empowered by GPT-4V \cite{achiam2023gpt} to generate object captions $c^t_i$ for the image $I^t_{R_i}$. Then, we leverage a visual detector $\text{VD}(\cdot)$ named GroundingDINO \cite{liu2023grounding}, to generate the bounding boxes $obb^t_i$ for captioned objects by
\begin{equation}
\setlength\abovedisplayskip{3pt}%shrink space
\setlength\belowdisplayskip{3pt}
\label{step_pos}
obb_i=\text{VD}(c^t_i, I^t_{R_i}), c^t_i = \text{CAP}(I^t_{R_i}).
\end{equation}
Finally, the bounding boxes are tokenized by a visual tokenizer $\text{VT}(\cdot)$, which is then fed into a segmentation model $\text{SEG}(\cdot)$ \cite{kirillov2023segment} to generate fine-grained semantic masks $I^t_{S_i}$ for objects as follows:
\begin{equation}
\setlength\abovedisplayskip{3pt}%shrink space
\setlength\belowdisplayskip{3pt}
\label{step_pos}
I^t_{S_i} = \text{SEG}(\text{VT}(obb_i), I^t_{R_i}).
\end{equation}

\noindent \textbf{Scene Spatial Perception}
Considering that ego-centric views suffer from perspective overlap \cite{an2023bevbert} and fail to capture 3D spatial relationships \cite{gao2024aerial}, we construct a 3D point map by projecting segmentation masks of observations into metric 3D space. Leveraging the RGB-D sensor's depth map $I^t_{D_i}$ and agent's pose $(R, T)$, a geometric projector (GP) transforms each segmented pixel $p_{ik}=(u, v) \in I^t_{S_i}$ labeled with caption $c^t_{ik}$ into a 3D point $P_{ik}$ via:
\begin{equation}
\setlength\abovedisplayskip{3pt}%shrink space
\setlength\belowdisplayskip{3pt}
\label{step_pos}
% P = R \cdot Z\cdot K^{-1} \cdot p + T, Z=I_D(u, v),
P_{ik} =R \cdot I_D(u, v)\cdot K^{-1} \cdot p + T,
\end{equation}
where $K$ is the intrinsic matrix of the camera, while $R\in\textit{SO}(3)$ and $T \in \mathbb{R}^3$  represent the agent's instantaneous orientation and position in world coordinates. Mapping the object caption from 2D masks to 3D point cloud, a local semantic point cloud $\{(P_{ik}, c^t_{ik})|i=1,\dots,n, k=1,\dots, m\}$ is constructed, where $n$ is the number of panoramic images and $m$ is the number of pixels.

\subsection{Hierarchical Semantic Planning Module}
\label{LLM planner}

\subsubsection{Landmark-level Planning}
Since aerial VLN tasks typically involve long-range decision-making \cite{liu2023aerialvln, chen2019touchdown}, directly assigning the entire navigation task to the agent can hinder accurate alignment with the instructions and task progress tracking. A more effective approach is to decompose the task into a sequence of manageable sub-goals. By addressing these sub-goals step by step, the agent can progressively reach the final destination. To achieve this, we propose a landmark-level planner driven by LLM \cite{achiam2023gpt} to parse free-form instructions $T$ and extract a sequence of landmark phases $L$ along the path through prompt engineering.
These landmarks act as sub-goals for the agent. 
We present a simple prompt as follows (more details in Appendix A):
\tcbset{colback=gray!30, colframe=gray!50, boxrule=0pt, arc=0pt, outer arc=0pt}
\begin{tcolorbox}
\textit{You need to extract a landmark sequence from the given instruction. The sequence order should be consistent with their appearance order on the path.
}
\end{tcolorbox}

\subsubsection{Object-level Planning}
After landmark-level planning and obtaining a sequence of sub-goals, the object-level planner $\text{OP}(\cdot)$ employs the LLM to further decompose these sub-goals into more achievable steps for the agent. The key idea is to leverage the commonsense knowledge of the LLM to reason for the visible object region most pertinent to the invisible sub-goal in the current panorama. This region is referred to as the object region of interest (OROI) in this paper. For example, if the agent only sees the buildings and a road in the current view while its sub-goal is the traffic light, by commonsense reasoning, the next OROI it should go is the road. We design a prompt that comprises the original navigation instruction $T$, scene object captions $c^t$, and current sub-goals $L_i$ for $\text{OP}(\cdot)$ to reason for OROI $c^t_{OROI}$, which is given by:
\begin{equation}
\setlength\abovedisplayskip{3pt}%shrink space
\setlength\belowdisplayskip{3pt}
\label{step_pos}
c^t_{OROI} = \text{OP}(T, L_i, c^t),
\end{equation}
Its template is (more details in Appendix~\ref{apdx:prompt_engineering}):

\begin{tcolorbox}
\textit{
The navigation instruction is: .... Your next navigation subgoal is: ... Objects or areas you observed: ... \\
\\
Based on the instruction, next subgoal, and observation, list 3 objects most pertinent to the subgoal or you will probably go next from your [Observed Objects]. Output should be in descending order of probability. }
\end{tcolorbox}
We select the OROI with the highest possibility given by LLM to the next landmark as the next waypoint for the agent.

\subsubsection{Motion-level Planning}
\label{waypoint planner}
Motion-level planning is responsible for translating the output of high-level planning modules into reachable waypoints and executable actions for the agent. Given a reasoned $c^t_{OROI}$, the motion-level planner first determines corresponding points $\{(P_k, c^t_k)|c^t_k == c^t_{OROI}\}$ from the semantic point cloud in \textsection \ref{scene perception module} and compute the next waypoint by averaging the coordinates of selected points. Then, the planner decomposes the path to the waypoint into a sequence of executable actions for the agent.

If the agent has reached a location close to the memory graph, the motion planner will directly use the memory graph to predict the agent's future actions, which is introduced in the next section. 

\subsection{Global Memory Module}
\label{memory module}
Since the agent sometimes revisits the target or landmarks, we designed a global memory module with a memory graph that stores historical trajectories, which helps to reduce the action space in motion planning and improves navigation robustness. Different from prior works that rely on predefined memory graphs \cite{chen2021topological, chen2022think} or 2D grid maps \cite{wang2023gridmm} lacking height information, our approach constructs a 3D topological memory graph from the agent’s navigation history.

\noindent \textbf{Memory Graph Construction} 
Each historical trajectory $H_i$ can be represented as a topological graph $G_i(N_i, E_i)$ whose nodes $N_i$ encapsulate both the coordinates of the traversed waypoints and their panoramic observations, and edges $E_i$ are weighted by the distance between adjacent waypoints. The memory graph $M$ is constructed by merging all the historical trajectory graphs, given by:
\begin{equation}
\setlength\abovedisplayskip{3pt}%shrink space
\setlength\belowdisplayskip{3pt}
\label{eq:mg_build}
\begin{aligned}
M &= G(N, E), \\
N &= N_1\cup \cdots \cup N_d, \\
E &= E_1\cup\cdots \cup E_d,
\end{aligned}
\end{equation}
where $d$ is the number of historical trajectories.

\noindent \textbf{Memory Graph Update} 
The memory graph is updated progressively by merging newly generated historical trajectory graph $G_{hist}$. The merging process is similar to Equation~\ref{eq:mg_build} merging the nodes and edges of two graphs. In addition, it will generate new edges if $M$ and $G_{hist}$ are connective. 
We calculate the distance between every pair of nodes in the two graphs. If the distance between any pair of nodes is less than a threshold $H$=15m, it is inferred that these nodes are adjacent and a new edge is added to the merged graph. 
Note that the memory graph only merges trajectories that successfully navigate to the target, ensuring the validity of the waypoints in the memory graph.

\noindent \textbf{Memory Graph Search for Motion-level Planning}
When the agent reaches a waypoint in the memory graph, the agent directly leverages the graph to determine a path and action sequence to fulfill the remaining sub-goals from the landmark-level planner. Given a sequence of remaining sub-goals $L^{(r)}=\{l_1, \dots, l_r\}$ represented by landmark phases, our target is to find a path $V^*=\{v_1, \dots, v_d\} \subseteq N$ with the highest possibility of traversing $L_r$ in order. Note that each node $N_i$ in the graph contains visual observation $o_{v_i}$ of the surroundings. Therefore, the possibility of traversing landmark $l_j$ by a node $N_i$ can be formulated as $P(l_j|o_i)$.
And the objective function can be formulated as:
\begin{equation}
\setlength\abovedisplayskip{3pt}%shrink space
\setlength\belowdisplayskip{3pt}
\label{eq:mg_search}
\begin{aligned}
V^* = \max_{V} \prod_{k=1, 1\leq m_1 < \cdots<m_r \leq d}^{r} P(l_k|o_{v_{m_k}}).
\end{aligned}
\end{equation}
We leverage the graph search algorithm in LM-Nav \cite{shah2023lm} to solve this problem, which is more detailed in Alg. 1 in Appendix \ref{apdx:graph-search}. Once $V^*$ is obtained, the agent decomposes it into an executable action sequence to reach the target.
To reduce the high complexity and inefficiency of searching over the full global memory graph, the agent extracts a subgraph for navigation. The agent first selects nodes and edges within a radius $R$ to form a spherical subgraph. It then computes the semantic relevance between subgraph nodes and the landmarks, and applies non-maximum suppression to prune redundant nodes. This results in a sparse memory subgraph for efficient and accurate navigation. More details in Appendix~\ref{apdx:graph_pruning}.

To conclude, with the two specially designed perception and planning modules, along with the memory module, the aforementioned key challenges of the aerial VLN are addressed one by one.

\begin{table*}
  \caption{Overall performance comparisons on AirVLN-S.}
  \label{tab:overall-performance-1}
  \centering
  \resizebox{0.9\linewidth}{!}{
  \begin{tabular}{lcccccccccc}
    \toprule
    \multirow{2}*{Method} & \multicolumn{5}{c}{Validataion Seen} & \multicolumn{5}{c}{Validation Unseen}            \\
    \cmidrule(r){2-6}
    \cmidrule(r){7-11}
    ~     & SR$\uparrow$  & SPL$\uparrow$  & OSR$\uparrow$ & SDTW$\uparrow$ & NE$\downarrow$ & SR$\uparrow$  & SPL$\uparrow$  & OSR$\uparrow$ & SDTW$\uparrow$ & NE$\downarrow$ \\
    \midrule
    RS & 0.0  & 0.0 & 0.0 & 0.0  & 109.6 & 0.0 & 0.0  & 0.0 & 0.0 & 149.7  \\
    AC & 0.9 & - & 5.7 & 0.3 & 213.8 & 0.2 & - & 1.1 & 0.3 & 237.6 \\ 
    \hdashline
    LingUNet & 0.6 & - & 6.9 & 0.2 & 383.8 & 0.4 & - & 3.6 & 0.9 & 368.4 \\
    Seq2seq & 4.8 & - & 19.8& 1.6 & 146.0 & 2.3 & - & 11.7 & 0.7 & 218.9 \\
    CMA & 3.0 & - & 23.2 & 0.6 & 121.0 & 3.2 & - & 16.0 &1.1&172.1 \\
    \hdashline
    NavGPT & 0.0 & 0.0 & 0.0 & 0.0 & 163.5 & 0.0 & 0.0 & 0.0 & 0.0 & 82.1 \\
    MapGPT & 2.1 & 1.5 & 4.7 & 0.8 & 124.9 & 0.0 & 0.0 & 0.0 & 0.0 & 107.0 \\
    VELMA  & 0.0 & 0.0 & 0.0 & 0.0 & 150.5 & 0.0 & 0.0 & 0.0 & 0.0 & 117.4 \\
    LM-Nav & 12.5 & 9.4 & 28.5 & 4.6 & 81.1 & 10.4 & 9.3 & 33.9 & 4.7 & 60.3 \\
    STMR & 12.6 & - & \textbf{31.6} & - & 96.3 & 10.8 & - & 23.0  & - & 119.5 \\
    \hdashline
    CityNavAgent  & \textbf{13.9}  & \textbf{10.2}  &   30.2 & \textbf{5.1}  & \textbf{80.8}  &  \textbf{11.7} & \textbf{9.9}  & \textbf{35.2}  &  \textbf{5.0} & \textbf{60.2} \\
    \bottomrule
  \end{tabular}
  }
\end{table*}

\section{Experiments}
\label{others}

\vspace{-0.3cm}
\subsection{Experimental Setup}
\label{exp-setup}
\textbf{Datasets} 
We evaluate CityNavAgent on a novel aerial VLN benchmark named AirVLN-S provided by Liu et al.\cite{liu2023aerialvln}. The benchmark is collected in Unreal Engine 4 to mimic real-world urban environments. It contains 25 different city-level scenes including downtown cities, factories, parks, and villages, with more than 870 different kinds of urban objects. It also consists of 3,916 flying paths collected by experienced UAV pilots. 

While AirVLN provides a valuable benchmark, it suffers from ambiguous landmark references in its relatively coarse-grained instructions. This lack of explicit spatial grounding (e.g., "going straight to the buildings") makes it challenging to systematically assess the agent’s performance at following each part of the instruction. To address this limitation, we follow the similar idea of \cite{hong2020sub} to enrich the original instruction with sub-instructions and their corresponding paths. We collect 101 fine-grained instruction-path pairs in 10 scenes from AirVLN to construct an instruction-enriched aerial VLN benchmark, named AirVLN-\textbf{E}nriched. The details are in Appendix \ref{apdx:refine-dataset}.

\noindent \textbf{Metrics} 
Following the same metrics used in AirVLN, we report and compare Success Rate (SR), Oracle Success Rate (OSR), Navigation Error (NE), SR weighted by Normalized Dynamic Warping (SDTD) and SR weighted by Path Length (SPL) of tested methods. The task is successfully completed if NE is within 20 meters.

\noindent \textbf{Implmentation Details}
We take the training samples as the historical tasks and initialize the memory graph by their trajectories. The memory graph remains accessible to the agent throughout the evaluation.
In each test case, the agent is spawned at a random location in the scene. It first follows the instruction to explore the environment, and upon reaching the memory graph, it leverages the graph to complete the rest of the navigation path. It has six low-level actions: \textit{Forward}, \textit{Turn Left}, \textit{Turn Right}, \textit{Ascend}, \textit{Descend}, \textit{Stop}. The number of total action steps is counted based on low-level actions. If the agent requires $n$ low-level actions to reach the next waypoint, the action count increases by $n$. The agent stops when it either triggers the stop action or exceeds the maximum action steps.

\noindent \textbf{Baselines} 
We choose three mainstream types of continuous VLN baselines. 
\begin{itemize}[leftmargin=*,partopsep=0pt,topsep=0pt]
\setlength{\itemsep}{0pt}
\setlength{\parsep}{0pt}
\setlength{\parskip}{0pt}
    \item \textbf{Statistical-based Methods.} We use random sample (RS) that agents uniformly select an action from the action space at each step and action sample (AC) that agents sample actions according to the action distribution of the dataset as our baselines.  
    \item \textbf{Learning-based Methods.} We choose classic learning-based methdods {Seq2Seq} \cite{anderson2018vision}, {CMA} \cite{krantz2020beyond}, and {LingUNet} \cite{misra2018mapping} as our baselines.
    \item \textbf{Zero-shot LLM-based methods.} We use SOTA outdoor VLN methods VELMA \cite{schumann2024velma}, LM-Nav \cite{shah2023lm}, and STMR \cite{gao2024aerial} as baselines. To validate the effectiveness of indoor VLN methods, we also evaluate SOTA indoor VLN methods: NavGPT \cite{zhou2024navgpt} and MapGPT \cite{chen2024mapgpt}.
\end{itemize}

\begin{table}
  \caption{Overall performance comparisons AirVLN-E.}
  \label{tab:overall-performance-2}
  \centering
  \resizebox{0.8\linewidth}{!}{
  \begin{tabular}{lccc}
    \toprule
    Methods & SR$\uparrow$ & SPL$\uparrow$  & NE$\downarrow$ \\
    \midrule
    RS & 0.0 & 0.0 & 129.6   \\
    AC & 0.0 & 0.0 & 290.4 \\ 
    \hdashline
    Seq2seq & 0.0&0.0&398.5\\
    CMA &0.0&0.0& 278.3\\
    \hdashline
    NavGPT & 0.0 & 0.0 & 127.2 \\
    MapGPT & 3.3 & 1.5 & 133.7 \\
    VELMA  & 0.0 & 0.0 & 138.0  \\
    LM-Nav & 23.6 & 19.2 &119.4 \\
    \hdashline
    CityNavAgent  & \textbf{28.3} & \textbf{23.5} & \textbf{95.1} \\
    \bottomrule
  \end{tabular}
  }
\end{table}

\subsection{Overall Performance}
In Table \ref{tab:overall-performance-1} and Table~\ref{tab:overall-performance-2}, we report the overall performance of CityNavAgent and baselines on the two aerial VLN benchmark. From these results, we have the following observations:
\begin{itemize}[leftmargin=*,partopsep=0pt,topsep=0pt]
\setlength{\itemsep}{0pt}
\setlength{\parsep}{0pt}
\setlength{\parskip}{0pt}
\item \textbf{CityNavAgent significantly outperforms previous SOTAs}. 1) Statistical-based methods have the worst performance indicating aerial VLN requires stronger planning capacity rather than random guess. 2) Learning-based methods that predict agent's action directly also have relatively poor performance with SR less than 5\%, which can be explained by the complex action space for long-range navigations. 3) Indoor LLM-based methods suffer significantly performance drop while outdoor LLM-based methods have better performance. 4) Compared to these baselines, CityNavAgent outperforms the best of them by 1.3\%, 0.8\%, 0.5\%, and 16.1\% in SR, SPL, SDTW and NE for validation seen dataset and by 0.9\%, 0.6\%, 1.3\%, and 0.2\% in SR, SPL, SDTW and NE for validation unseen dataset. It demonstrates that the semantic hierarchical planning and memory graph-based motion planning improve the agent's long-range navigation capacity.

\item \textbf{CityNavAgent has better instruction-following performance.} We can
observe that our proposed CityNavAgent achieves the highest SPL and STDW, which outperforms the best of baselines by 0.8\% and 0.5\%, respectively. We explain the cause of this result as: 1) the HSPM decomposes the original long-distance navigation task into shorter sub-navigation tasks, reducing the planning difficulty. 2) memory graph-based motion planning further guides the agent to traverse the decomposed landmark sequence.

\item \textbf{Enriched Instructions Promote Navigation Performance.} The performance of MapGPT, LM-Nav, and CityNavAgent on AirVLN-E is better than on AirVLN-S. The highest improvements in SR and SPL are 14.4\% and 13.3\%, respectively, indicating that the enriched instructions provide clearer landmarks, helping the agent follow the path to the target more effectively.

\end{itemize}

\subsection{Ablation Study}
\textbf{Effect of semantic map-based exploration.}
To evaluate the effectiveness of semantic map-based waypoint prediction, we substitute this module with a random walk strategy. As shown in Table \ref{ablations}, the agent without the semantic map (second row) suffers a 4.7\% and 4.3\% drop in SR and SPL, and a 25.6\% increase on NE with CityNavAgent (last row). This result reveals that the semantic map extracts structured environmental information, facilitating the LLM in commonsense reasoning so that the agent navigates to the region or objects that are more relevant to the navigation task. Consequently, the accuracy and efficiency of the navigation is improved.  

\begin{table}[t]
\footnotesize
\vspace{-0.5cm}
  \caption{Effectiveness of different modules in CityNavAgent. MG and SM represent the memory graph and semantic map, respectively.}
  \label{ablations}
  \centering
  \resizebox{\linewidth}{!}{
  \begin{tabular}{l|ccc}
  \hline
  \hline
Modules & SR$\uparrow$  & SPL$\uparrow$  & NE$\downarrow$ \\
    \hline
    w/o MG & 11.7 & 9.1 & 206.1 \\
    w/o SM & 23.6 & 19.2 & 119.4 \\
    w/ LLaVA-7B & 1.7 & 1.4 & 125.7 \\ 
    w/ GPT-3.5 & 23.3 & 16.1 & 98.9 \\
    w/ GPT-4V (ours) & 28.3 & 23.5 & 95.1 \\
    \hline
    \hline
  \end{tabular}
  }
  % \vspace{-0.4cm}
\end{table}

\begin{table}[t]
    \centering
    \caption{Comparison of waypoint predictor under different scenarios.}
    \label{waypoint stat}
    \resizebox{\linewidth}{!}{
    \begin{tabular}{c|l|cccc}
    \hline
    ~ & Inputs & $|\triangle|$ & $d_{rel}\downarrow$ & $d_{C}\downarrow$ & $d_{H}\downarrow$ \\
    \hline
    \multirow{3}*{Outdoor} & RGBD   & 1.66   & 0.88   & 6.46   & 5.16   \\
                          & RGB   & 1.59   & 0.88   & \textbf{6.40}   & \textbf{5.15}  \\
                          & Depth  & 1.60  & 0.88  & 6.48  & 5.26  \\
    \hdashline
    \multirow{3}*{Indoor}  & RGBD & 1.40 & - & 1.05 & 2.01 \\
                           & RGB & 1.38 & - & 1.08 &  2.03 \\
                           & Depth & 1.39 & - & \textbf{1.04} & \textbf{2.01} \\
    \hline
    \end{tabular}
    }
\end{table}

\textbf{Effect of memory graph-based exploitation.}
In this case, we omit the global memory module during the navigation and replace the graph search algorithm with the random walk strategy. Presented in the first row in Table \ref{ablations}, the lack of memory graph results in a 16.6\%, 14.4\% decrease in SR, SPL, and a 116.7\% increase in NE over CityNavAgent. Moreover, the memory graph has a more noticeable impact on the agent's navigation performance compared with the semantic map. This indicates that the memory graph effectively prevents the agent from falling into dead ends or engaging in blind exploration in long-distance outdoor navigation scenarios, thereby ensuring the stability of navigation performance.

\begin{figure}
  \centering
  \setlength{\abovecaptionskip}{0 pt}
  \includegraphics[width=0.8\linewidth]{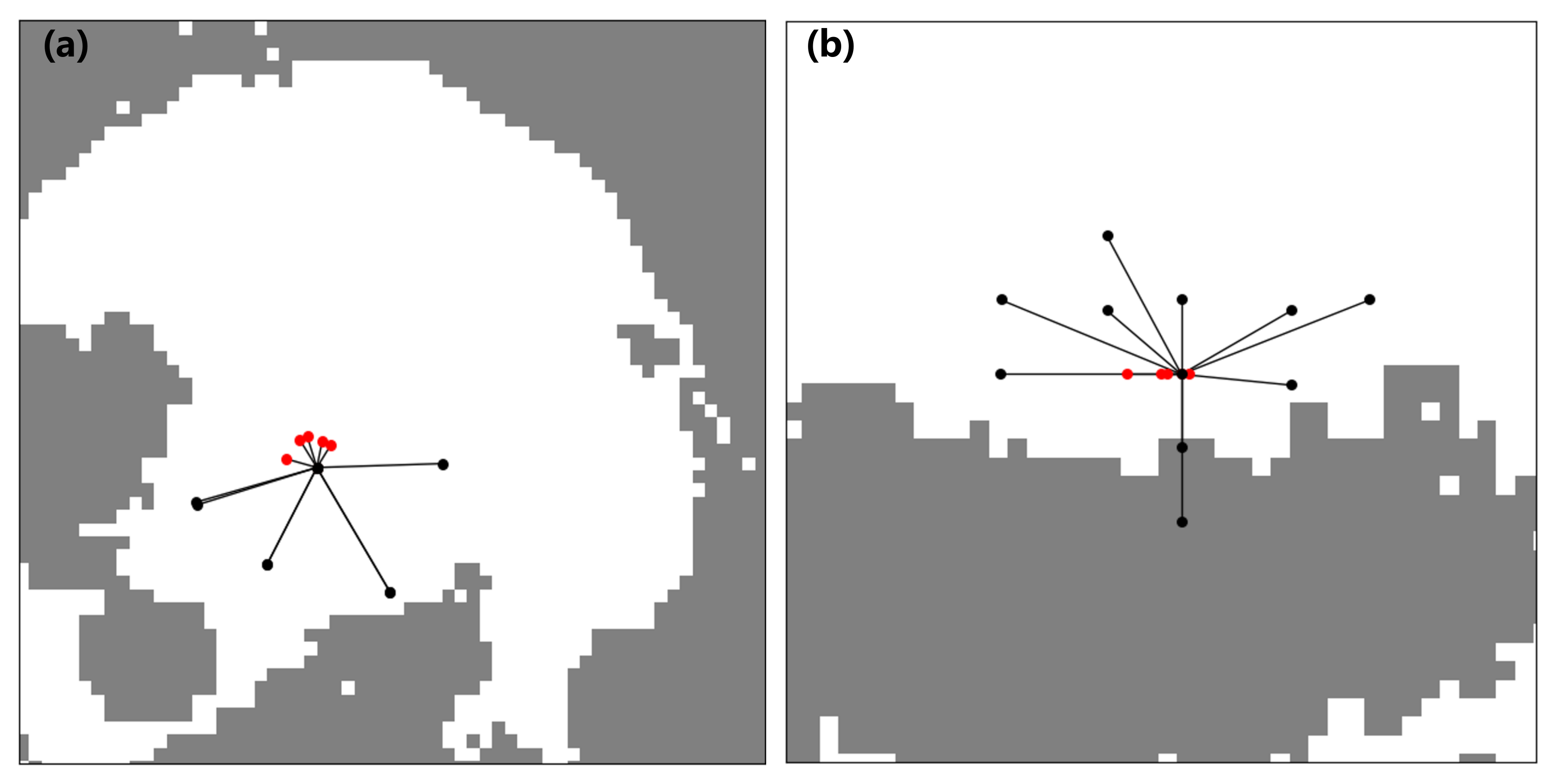}
  \caption{The qualitative result of CWP. From left to right are the top-down view and the front-facing view. \textbf{\textit{\textcolor[rgb]{1, 0.0, 0.00}{Red}}} and \textbf{\textit{\textcolor[rgb]{0, 0, 0}{black}}} circles denote predicted waypoints and reference waypoints on real trajectories.}
  \label{rq3}
  \vspace{-0.5cm} 
\end{figure}

\textbf{Effect of different LLMs.}
We also evaluate the effectiveness of different LLMs for commonsense reasoning in object-level planning (\textsection \ref{LLM planner}). Agent with LLaVA-7B as the object-level planner achieves the poorest performance, which is mainly due to the perception hallucination and unstructured output formats. Although CityNavAgent with GPT-3.5 demonstrates a competitive performance, replacing GPT-3.5 with GPT-4V which has enhanced reasoning capability results in further performance improvement, \textit{e.g.}, 5.0\% and 7.4\% increases in SR and SPL, respectively. We attribute this improvement to the fact that GPT-4V has a lower hallucination rate and stronger reasoning ability. Thus, it generates more contextually appropriate responses based on the semantic map and navigation instruction to facilitate the agent in exploring areas most relevant to the target.

\subsection{Effectiveness of Indoor Waypoint Prediction}
To evaluate the effectiveness of the waypoint prediction method CWP \cite{hong2022bridging} in previous continuous VLN methods \cite{an2024etpnav, koh2021pathdreamer, wang2023dreamwalker,wang2023gridmm, krantz2022sim, wang2024lookahead}, we compare the predicted waypoint with target waypoints in outdoor environments. 
We apply waypoint metrics \cite{hong2022bridging} to assess the quality of predicted waypoints. $|\triangle|$ measures the difference in the number of target waypoints and predicted waypoints. $d_{rel}$ measures the ratio of average waypoint distance. $d_{C}$ and $d_{H}$ are the Chamfer distance and the Hausdorff distance, respectively. As depicted in Table \ref{waypoint stat}, CWP achieves the best performance in indoor environments with 1.04 $d_C$ and 2.01 $d_H$ while in outdoor environments with 6.4 $d_C$ and 5.15 $d_H$. 
This result indicates that although the predicted indoor waypoints by CWP are close to the indoor target waypoints, predicted outdoor waypoints are far from outdoor target waypoints, which is illustrated intuitively in Figure \ref{rq3}(a). We attribute this to the scale difference between indoor and outdoor environments. Besides, the dimensional difference is another negative factor for CWP. Depicted in Figure \ref{rq3}(b), CWP only predicts waypoints in 2D space and cannot be applied to open urban environments.

\begin{figure}[t]
  \centering
  \setlength{\abovecaptionskip}{-0pt}
  \includegraphics[width=0.8\linewidth]{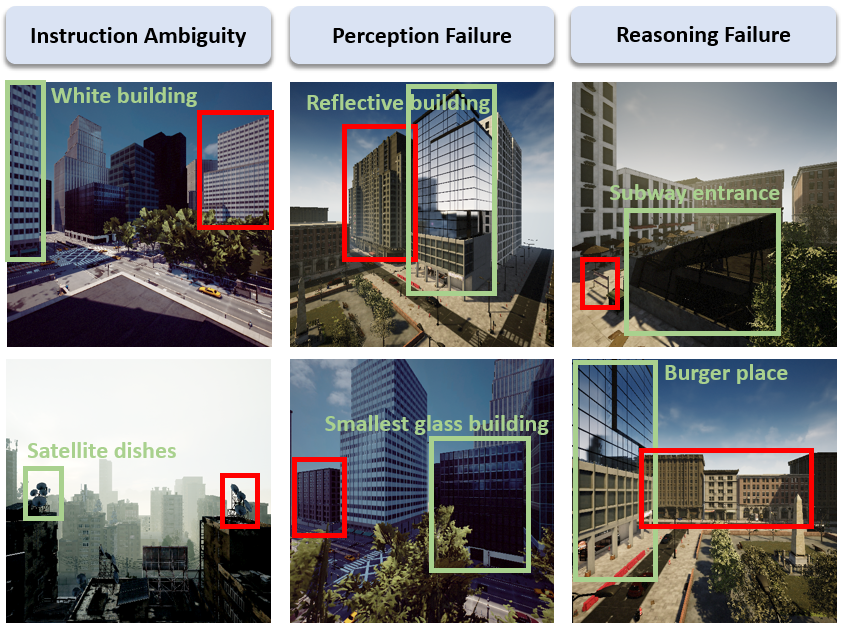}
  \caption{The qualitative result of failure cases. The \textbf{\textit{\textcolor[rgb]{0.66, 0.66, 0.56}{green}}} captions and bounding boxes are the referred landmarks in instructions. The \textbf{\textit{\textcolor[rgb]{1, 0, 0}{red}}} bounding boxes are the misreferred landmarks due to three failure reasons.}
  \label{fig:failure_case}
  \vspace{-0.3cm} 
\end{figure}

\subsection{Case Analysis}

In this section, we analyze the failure cases across different datasets and categorize them into three types: 1) Instruction Ambiguity: The navigation instruction lacks clear landmarks, or there are many similar landmarks within the same scene, making it difficult for the agent to accurately refer to the landmark mentioned in the instruction. 2) Perception Failure: Despite the strong object recognition and detection capabilities of our open-vocabulary perception module, outdoor scenes still present many edge cases, leading to incorrect identification of landmarks referenced in the instruction. 3) Reasoning Module: During the hierarchical planning process, the object-level planner may encounter reasoning errors when attempting to infer the location of OROI. This can happen when there is insufficient semantic connection between the objects in the scene and the referenced landmarks, resulting in incorrect OROI reasoning.
The visualization results of these failure cases are shown in Figure \ref{fig:failure_case}. More qualitative results are in Appendix~\ref{subsec:failure_distribution}.

\vspace{-0.2cm}
\section{Conclusion}
\vspace{-0.2cm}
In this paper, we approach the problem of zero-shot vision-language navigation by proposing an embodied aerial agent, CityNavAgent, which leverages the pre-trained knowledge in large foundation models and historical experience to deal with long-term navigation in urban spaces. 
The experimental results illustrate the efficacy and robustness of our method from different perspectives.

\section*{Limitations}
\vspace{-0.2cm}
One limitation of our work is that the whole system has not been deployed on a real drone. Though our methods achieve promising results in simulated outdoor environments, low-level motion control problems such as self-pose estimation, control latency, and control errors are not considered in our work. The second is that the agent lacks a backtracking mechanism. CityNavAgent so far only relies on the sub-goal decomposition to track the navigation path.

\bibliography{ref}

\clearpage
\appendix
\section{Appendix}
\subsection{Prompt Engineering}
\label{apdx:prompt_engineering}
In this part, we discuss the prompt used in this work. 
<> denotes the content needs to be filled in the prompting process.

\textbf{LLM Prompt for High-level Planner}
The complete prompt is:

\begin{tcolorbox}[title=Task Planning Prompt]
\textit{You are a navigation aircraft, and now you need to navigate to a specified location according to a natural language instruction. You need to extract a landmark sequence from the instruction. The sequence order should be consistent with their appearance on the path. Your output should be in JSON format and must contain two fields: "Landmark sequence" and "Thought." "Landmark sequence" is your thinking result comprised of landmark phrases in the instruction. "Thought" is your thinking process.  
The instruction is <
"First, you need to find a stop sign. Then take left and right and continue until you reach a square with a tree. Continue first straight, then right, until you find a white truck. The final destination is a white building." >}
\end{tcolorbox}

The example output is: 

\begin{tcolorbox}
\textit{
"Landmark sequence": [ \\
    "stop sign", \\
    "square with a tree", \\
    "white truck", \\
    "white building", \\
    ] \\
  "Thought": "The instruction outlines a series of landmarks that need to be followed in order. The first landmark is a 'stop sign,' which I need to locate first. After finding the stop sign, I will navigate to a 'square with a tree' by taking left and right turns. Then, I will continue straight and turn right until I encounter a 'white truck.' Finally, my destination is a 'white building.' I have listed these landmarks in the order they appear in the navigation process."
}
\end{tcolorbox}

\textbf{The Prompt for Low-level Planner}
The complete prompt for the LLM commonsense reasoning in the low-level planner is:
\begin{tcolorbox}[title=Commonsense Reasoning Prompt]
\textit{
You are a drone and your task is navigating to the described target location! \\
\\ 
Navigation instruction: Start from the 'open' logo, fly forward and pass the yellow 'restaurant' logo in the front road. Then fly forward to the left, pass the circle grassland, and turn right. Finally, stop in front of the glass door with the trash bin and bench.\\
\\ 
Your next navigation subgoal: yellow 'restaurant' logo \\
\\
Objects or areas you observed: 
building with stairs, road, street lamp, street lamp, 'open' logo, building with stairs \\
Based on the instruction, next navigation subgoal, and observation, list 3 objects you will probably go next from your OBSERVED OBJECTS in descending order of probability.
}
\end{tcolorbox}

\begin{table*}
\footnotesize
\vspace{-0.5cm}
  \caption{Comparision of fine-grained dataset with AirVLN.}
  \label{fine-grained-stats}
  \centering
  \resizebox{\linewidth}{!}{
  \begin{tabular}{l|cccccccc}
  \hline
Dataset & Routes & Vocab & Instr. Len. & \# of Landmark & Traj. Len. & Traj. Len. (easy) & Traj. Len. (normal) & Traj. Len. (Hard) \\
    \hline
    AirVLN & 3,916 & 2.8k & 82 & - & 321.3 & - & - & -\\
    Refined & 101 & 0.4k & 39 & 4.1 & 156.1 & 104.3 & 147.9 & 235.8\\
    \hline
  \end{tabular}
  }
  % \vspace{-0.4cm}
\end{table*}

\subsection{Global Memory Module}
\subsubsection{Memory Graph Search Algorithm}
\label{apdx:graph-search}
The graph search problem is formulated as given a memory graph $G(N, E)$ and a sequence of landmark phrases $L=(\ell_1, \ell_2, \dots, \ell_n)$ extracted from the language instructions, the goal is to determine a sequence of waypoints $W=(w_0, w_1, \dots, w_m)$ that maximizes $P(r_L=1|W, L)$, where $r_L=1$ indicates that the sequence of the landmarks is traversed successfully. A scoring function $Q(i, w_k)$ is defined to represent the max probability of a path ending in $w_k$ that visited the landmarks $(\ell_1, \dots, \ell_i)$ and $P(r_L=1|W, L) = Q(n, W)$. Then, a graph search method integrated with the Dijkstra algorithm \cite{dijkstra2022note} is designed for calculating $W^*$. 

\begin{algorithm}[H]
\caption{Graph Search \cite{shah2023lm}}
\begin{algorithmic}[1]
\label{graph search}
\State \textbf{Input:} Landmarks $(\ell_1, \ell_2, \dots, \ell_n)$.
\State \textbf{Input:} Graph $G(N, E)$.
\State \textbf{Input:} Starting node $S$.
\State $\forall i=0,\dots,n, \forall w \in N, Q[i, w] \gets -\infty$
\State $Q[0, S] \gets 0$
\State \textsc{Dijkstra\_algorithm}$(G, Q[0, *])$
\State \For{$i \in 1, 2, \dots, n$}
    \State $\forall w \in W, Q[i, w] \gets Q[i-1, w] + \text{LLM}(w, \ell_i)$
    \State \textsc{Dijkstra\_algorithm}$(G, Q[i, *])$
    \EndFor
\State $destination \gets \arg\max(Q[n, *])$
\State \textbf{return} \textsc{backtrack}$(destination, Q[n, *])$
\end{algorithmic}
\end{algorithm}

\begin{figure*}[t]
  \centering
  \includegraphics[width=0.9\linewidth]{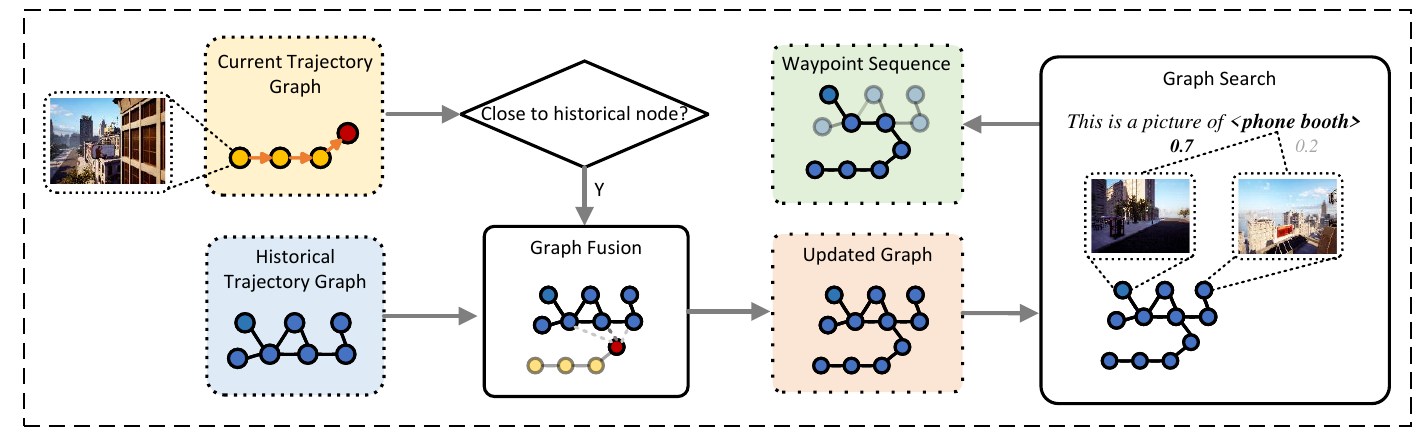}
  % \vspace{-0.2cm}
  \caption{ The illustration of the memory module. The agent stores its visual observation in the current trajectory graph. Once it reaches the node in the historical trajectory graph, the agent fuses these two graphs and searches for a path with the highest probability to the target. The probability is measured by a similarity score between the landmark phase in the instruction and visual observation stored in the node.}
    \vspace{-0.4cm}
  \label{waypoint_predict}
\end{figure*}

\subsubsection{Graph Pruning for Efficient Navigation}
\label{apdx:graph_pruning}
When the agent executes a new navigation task, it does not navigate through the entire global memory graph but instead on a subgraph selected from the global memory to improve the efficiency and accuracy of the graph searching algorithm. 

The subgraph is constructed in two stages. In the first stage, the agent samples all nodes and edges within a radius $R$ near the starting point on the global memory graph $G$ to form a subgraph $G_s$. $R$ is a hyperparameter, which is set to the average distance from the start point to the target in the training set in the experiment. Since $G_s$ still contains a significant amount of redundant nodes and edges, we further downsample it by a 3D non-maximum suppression (NMS) method in the second stage. First, we use CLIP to compute the matching score of each node $G_s$ with the landmarks in the instruction. Then, based on these matching scores, we apply NMS on $G_s$ to remove the nodes near local maxima and update the edges between the remaining nodes. Finally, we obtain a sparse memory subgraph $G_s$ for efficient and accurate navigation. Table~\ref{tab:graph_pruning} illustrates the node and edge counts of the global graphs and subgraphs. After graph pruning, the node counts of final subgraphs across all environments are reduced to less than 150, approximately half of the global map, thereby significantly improving the planning efficiency.

\subsubsection{Visualization of Memory Graph}
\label{apdx:mem_graph}

\begin{figure*}[t!]
  \centering
  \includegraphics[width=0.9\linewidth]{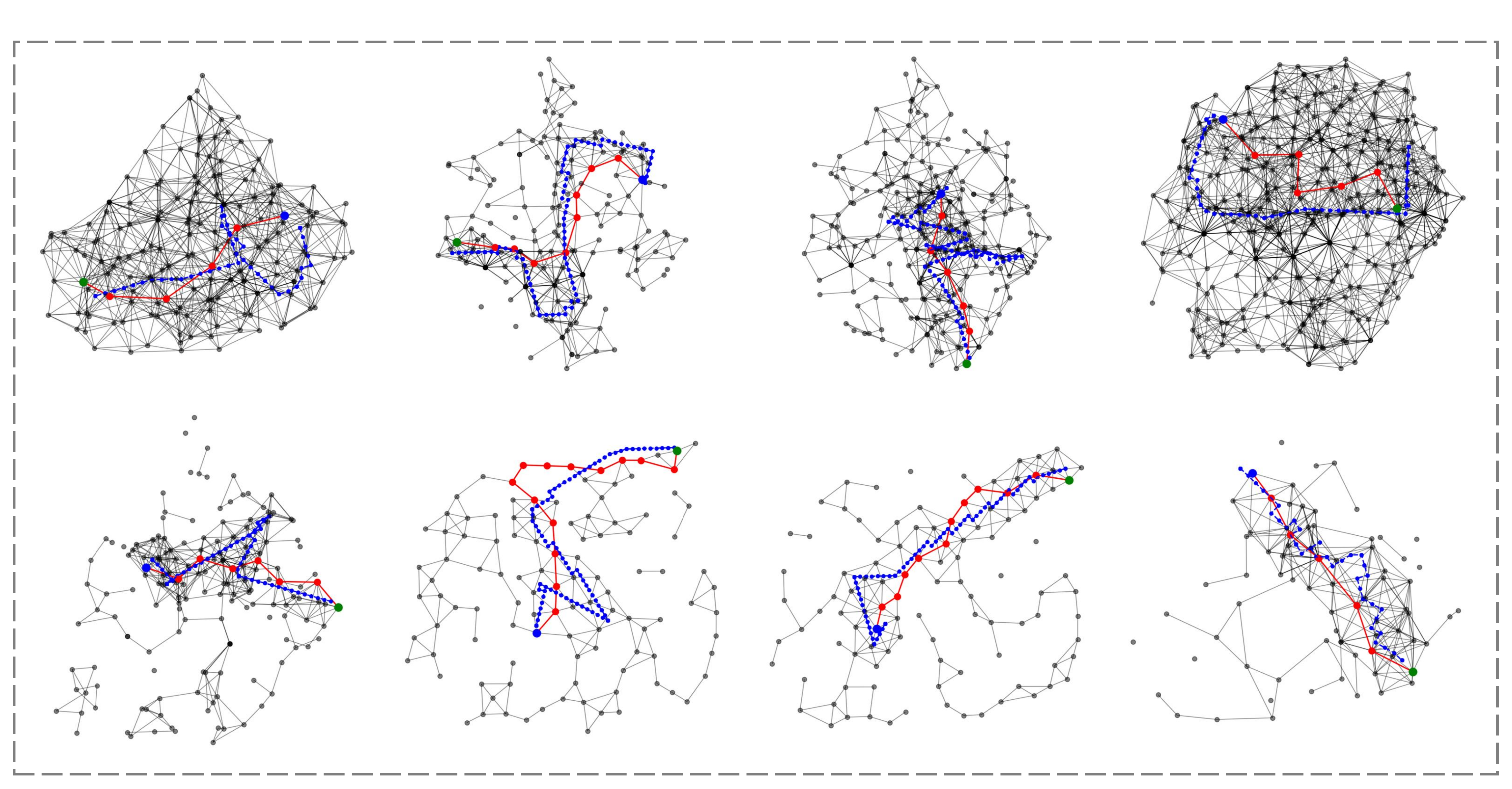}
  \caption{\small The memory graphs in different scenes. Each node in graphs stores the node's location and the agent's observation. The long distance between different historical trajectories results in the disconnection of memory graphs. \textit{\textcolor[rgb]{0, 0.0, 0.8}{Blue}} dot lines and \textit{\textcolor[rgb]{0.9, 0.0, 0.0}{red}} dot lines are ground truth and planned trajectories, respectively. The planned trajectories in the first row fail to follow the ground-truth trajectories while the last row have better instruction following the performance. }
  \vspace{-0.4cm}
  \label{fig:mem_graph_apdx}
\end{figure*}

The memory graph in different scenes is shown in Figure \ref{fig:mem_graph_apdx}.

\subsection{Point Clould Construction}
As shown in Fig.2, with the camera intrinsic matrix $K$ and agent's pose $(p, \alpha)$, pixels $p_o$ of the depth image of observation view can be projected to a 3D point cloud in the world coordinate system as $P_w^{\alpha^{\prime}} = R_{\alpha^{\prime}}\cdot Z \cdot K^{-1} \cdot p_{o} + p$, where $R_{\alpha^{\prime}}$ is the rotation matrix of observation view and $Z$ denotes depth values of pixels. The final point cloud $\mathcal{M}_s$ is given by $\mathcal{M}_s = P_w^{\alpha-90^{\circ}} \odot P_w^{\alpha-45^{\circ}} \odot P_w^{\alpha} \odot P_w^{\alpha + 45^{\circ}} \odot P_w^{\alpha+90^{\circ}}$. Note that RGB-D images in each observation view are well-aligned, meaning that the extracted semantic masks of RGB images have their counterparts in depth images as well as in the point cloud. To this end, a local map containing both semantic and spatial information is constructed.

\subsection{Details on Perception Module}
For all experiments, we employ GPT-4V \cite{achiam2023gpt} for object reasoning and landmark phase extraction. During the image grounding, a target is considered successfully detected if the bounding box's confidence score exceeds the threshold $\theta=0.4$. The agent is equipped with an aligned RGB-D camera with 512x512 resolution and $90^{\circ}$ field of view (FOV), capturing panoramic observations by rotating itself. The panoramic view directions are set at $p-90^{\circ}$, $p-45^{\circ}$, $p^{\circ}$, $p+45^{\circ}$, and $p+90^{\circ}$ where $p$ represents the agent's heading direction. The agent's low-level action space is (\textit{"move forward"}, \textit{"turn left"}, \textit{"turn right"}, \textit{"go up"}, \textit{"go down"}, and \textit{"stop"}). The moving step is 5 meters and each rotation turns the agent by $15^{\circ}$. The agent will receive its GPS location at each step.

\subsection{Fine-grained AirVLN Dataset}
\label{apdx:refine-dataset}
\begin{figure*}[t!]
  \centering
  \includegraphics[width=0.9\linewidth]{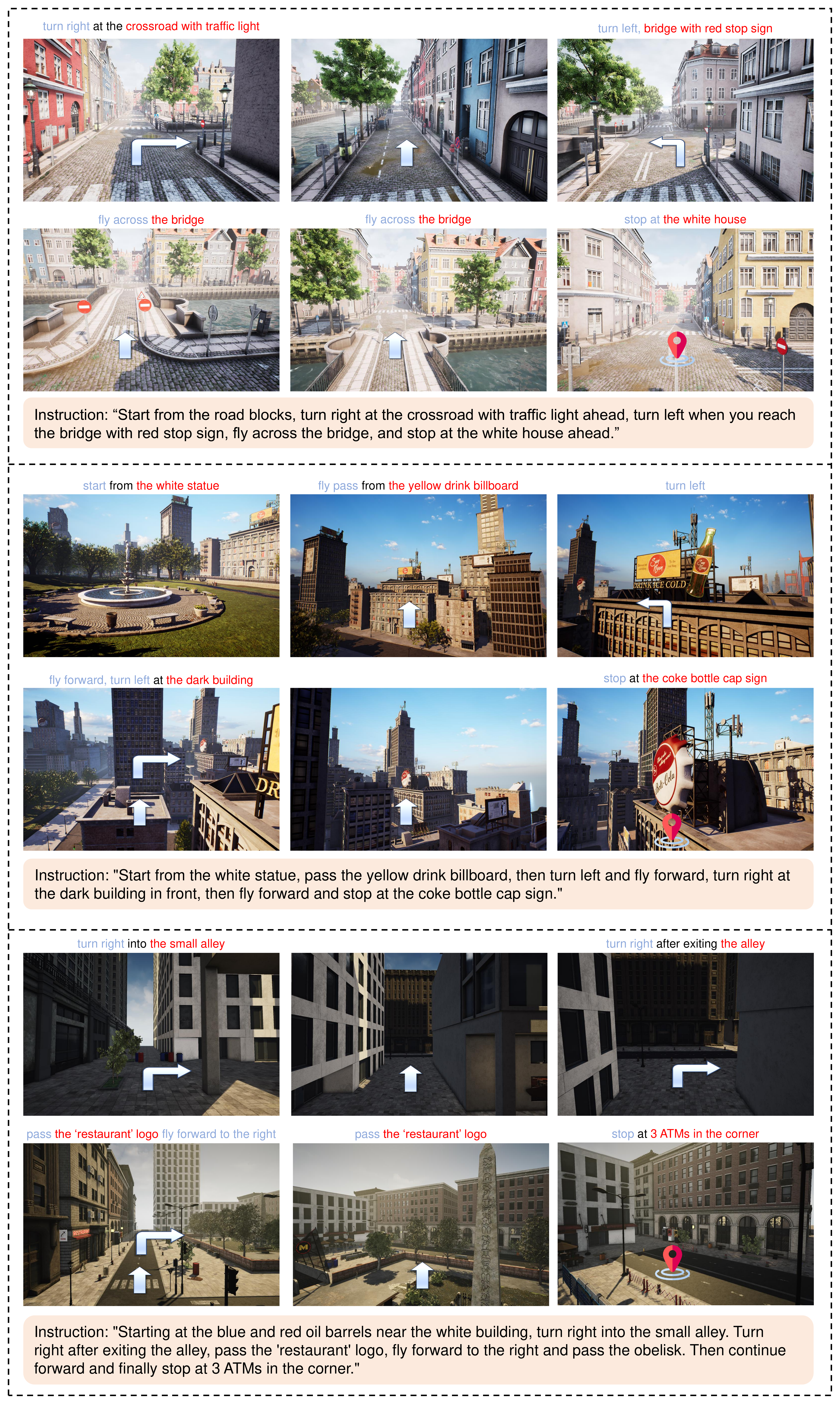}
 %   \vspace{-0.3cm}
  \caption{{The fine-grained AirVLN instruction and trajectory. The instructions and trajectories are well-aligned.}}
  \vspace{-0.4cm}
  \label{fine-grained}
\end{figure*}
We follow the similar idea of \cite{hong2020sub} to provide detailed descriptions of visual landmarks along the path and the agent’s actions, making instructions more specific. Fig. \ref{fine-grained} illustrated the sample of a fine-grained sample. The following are the comparison between a fine-grained instruction and an original AirVLN instruction:
\begin{tcolorbox}
\textit{\textbf{Instruction in AirVLN}: "turning left and going straight to the buildings and slight right turn. coming near the lake and turning right and going up to the building. coming down to the building and again going up and going straight. going top of the building and turning left and coming down to the building. roaming around a tower and searching each floor of apartment."}

\textit{\textbf{Fine-grained instruction}:Start from the blue billboard with 'leartes bank', fly forward, pass the 'GAS' sign, and turn left. Continue flying forward, passing the yellow billboard and the 'Americar' billboard ahead. Continue flying forward and left until you reach the blue billboard with 'leartes bank' and stop."}
\end{tcolorbox}

\textbf{Dataset statistics} The fine-grained samples are collected in 10 scenes from AirVLN. In each scene, 10 samples are collected. We further divide the fine-grained samples into three difficulty levels based on their trajectory length: easy tasks traverse two landmarks, normal tasks pass through three to four landmarks, and hard tasks involve navigating past five or more landmarks. The ratio of these three types of tasks is 1:3:1. The detailed statics are shown in Tab. \ref{fine-grained-stats}.

\begin{table}
    \centering
    \caption{Node and edge counts of different graphs. Env ID is the scene ID in AirVLN-s validation set.}
    \label{tab:graph_pruning}
    \resizebox{\linewidth}{!}{
    \begin{tabular}{lccc}
    \hline
    Env ID & Global Graph (Node/Edge) & Spherical Subgraph & Final Subgraph \\
    \hline
    2 & 177/587   & 110/353   & 71/135    \\
    3 & 44/242    & 30/150   & 25/91  \\
    5 & 395/2529  & 285/1821  & 130/413 \\
    8 & 48/203    & 29/116    & 24/70 \\
    10& 142/704   & 124/610   & 74/201 \\
    12& 76/285    & 56/206    & 40/94   \\
    14& 139/728   & 81/412    & 46/133   \\
    17& 39/126    & 21/61     & 18/41   \\
    \hline
    \end{tabular}
    }
\end{table}

\begin{table*}
  \caption{Overall performance comparisons AirVLN-E.}
  \label{tab:overall-performance-2-full}
  \centering
  \resizebox{\linewidth}{!}{
  \begin{tabular}{lcccccccccccc}
    \toprule
    \multirow{2}*{Method} & \multicolumn{3}{c}{Easy} & \multicolumn{3}{c}{Normal} & \multicolumn{3}{c}{Hard} & \multicolumn{3}{c}{Mean}             \\
    \cmidrule(r){2-4}
    \cmidrule(r){5-7}
    \cmidrule(r){8-10}
    \cmidrule(r){11-13}
    
    ~     & SR$\uparrow$      & SPL$\uparrow$  & NE$\downarrow$ & SR$\uparrow$      & SPL$\uparrow$  & NE$\downarrow$ & SR$\uparrow$      & SPL$\uparrow$  & NE$\downarrow$ & SR$\uparrow$ & SPL$\uparrow$  & NE$\downarrow$ \\
    \midrule
    RS & 0.0  & 0.0 & 85.6 & 0.0  & 0.0 & 127.9 & 0.0  & 0.0 & 164.9 & 0.0 & 0.0 & 129.6   \\
    AC & 0.0 & 0.0 & 242.2 & 0.0 & 0.0 & 315.6 & 0.0 & 0.0 & 263.2 & 0.0 & 0.0 & 290.4 \\ 
    \hdashline
    Seq2seq &0.0&0.0&201.1&0.0&0.0&359.4&0.0&0.0&713.3&0.0&0.0&398.5\\
    CMA &0.0&0.0&152.1&0.0&0.0&317.6&0.0&0.0&286.8&0.0&0.0& 278.3\\
    \hdashline
    NavGPT & 0.0 & 0.0 & 79.8 & 0.0 & 0.0 & 126.1 & 0.0 & 0.0 & 177.6 & 0.0 & 0.0 & 127.2 \\
    MapGPT & 0.0 & 0.0 & 97.5 & 5.6 & 2.9 & 135.2 & 0.0 & 0.0 & 165.5 & 3.3 & 1.5 & 133.7 \\
    VELMA     & 0.0 & 0.0  & 76.5 & 0.0 & 0.0  & 141.7 & 0.0 & 0.0  & 192.4 & 0.0 & 0.0 & 138.0     \\
    LM-Nav     & 15.4   & 13.7 & 123.1 & 22.2   & 18.1 & 124.3 & 33.3   & \textbf{28.1} & \textbf{114.2}  & 23.6 & 19.2 &119.4 \\
    \hdashline
    CityNavAgent  & \textbf{25.0}  & \textbf{21.3}  &   \textbf{74.7} & \textbf{27.8}  & \textbf{23.3}  &   \textbf{93.4} & \textbf{33.3}  & 26.3  &   121.5 & \textbf{28.3} & \textbf{23.5} & \textbf{95.1} \\
    \bottomrule
  \end{tabular}
  }
\end{table*}

\begin{figure*}[t!]
  \centering
  \includegraphics[width=0.8\linewidth]{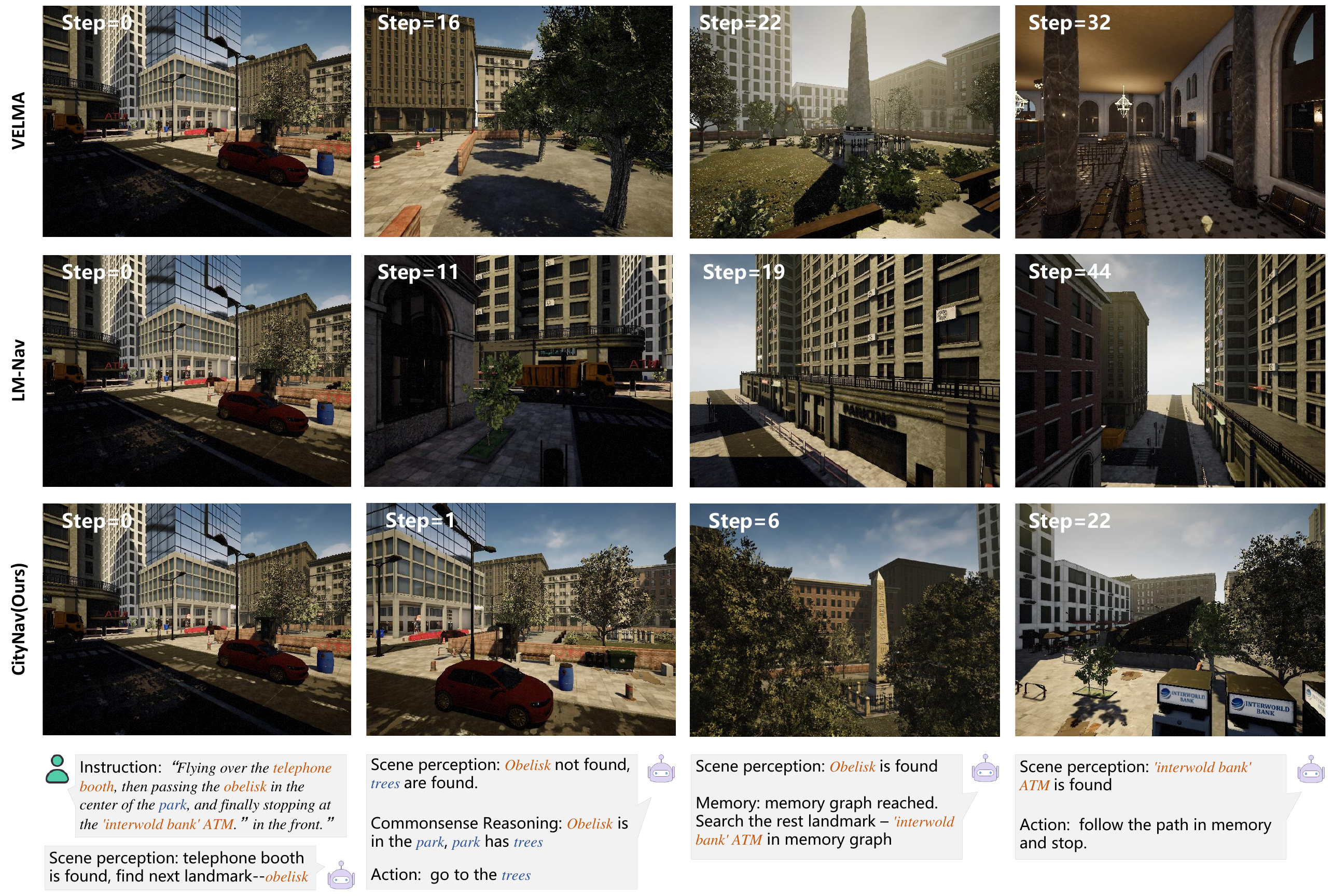}
  \caption{\small Qualitative result of the navigation process. The first three rows are the first-person view of VELMA, LM-Nav and CityNavAgent during the navigation. The last row is the reasoning process of CityNavAgent. \textit{\textcolor[rgb]{0.77, 0.35, 0.07}{Orange}} and \textit{\textcolor[rgb]{0.19,0.33,0.59}{blue}} represents the navigable landmarks in the instruction and common objects that semantically relevant to the landmarks, respectively.}
  \label{reason_process}
\end{figure*}

\subsection{More Experiment Results}
\subsubsection{Result on AirVLN-E of different difficulties}
In Table~\ref{tab:overall-performance-2-full}, we compare the overall performance of CityNavGPT with SOTA navigation methods, from which we make the following observations.
\begin{itemize}[leftmargin=*,partopsep=0pt,topsep=0pt]
\setlength{\itemsep}{0pt}
\setlength{\parsep}{0pt}
\setlength{\parskip}{0pt}
    \item \textbf{The vast action space exhibits great challenge for baselines.} Both the SR and SPL of statistical-based and learning-based methods are close to $0$, and the NE of these two methods are significantly high. This result illustrates that the long-term outdoor VLN involves an extremely large action space and even learning-based methods failed to traverse the whole space to find optimal paths.  
    \item \textbf{LLM fails to be a low-level action planner in continuous space.} Zero-shot LLM-based methods that leverage LLM for scene understanding and motion planning also have $0$ SR and high NE. We attribute this result to the fact that general-purpose LLMs like GPTs fail to reason for the low-level action due to the lack of domain-specific knowledge. Besides, directly predicting a long-term action sequence without intermediate checks will cause errors that accumulate along the path, leading to great deviation from the target. 
    \item \textbf{Our CityNavGPT method significantly outperforms previous SOTAs on all metrics}. Specifically, CityNavGPT improves SR by approximately 28.3\% and 4.7\% over VELMA and LM-Nav, respectively. This demonstrates the critical importance and navigation efficiency of semantic map-based waypoint planning for continuous outdoor navigation. Furthermore, in terms of SPL, our approach achieves improvements of 23.5\% and 2.5\% compared to VELMA and LM-Nav, respectively, indicating that CityNavGPT predicts navigation paths that are closer to the ground truth. The lowest error rate shows that even for those failed cases, our method still stops relatively close to the target. For easy and normal tasks, our method consistently surpasses the baseline by at least 5\% in SR and SPL. For hard tasks, our method still has a similar performance to the best method.
\end{itemize}

\subsubsection{More Results of Reasoning Process}

We illustrate the qualitative navigation process of CityNavAgent to further illustrate how the commonsense reasoning and memory graph work.
As depicted in Figure \ref{reason_process}, the agent is spawned at a random location with a navigation instruction. The agent has to explore the ordered landmarks in the instruction based on its visual observation. Thanks to its reasoning capabilities, the agent infers objects in its FOV that are semantically related to landmarks, even when those landmarks are not visually observed. In the given example, the agent tries to find the obelisk in the park which is currently invisible. Hinted by the instruction that the obelisk is in the park, CityNavAgent reasons that the trees probably appear in the park and decides to explore the areas near the trees while LM-Nav is gradually lost due to its lack of exploration ability (first three columns of Figure \ref{reason_process}). Once the agent reaches a place visited before, CityNavAgent leverages the memory graph to search for a path to the target while VELMA only relies on LLM for action planning and is trapped in an unfamiliar place (see the last columns of Figure \ref{reason_process}).  

\subsubsection{Failure Case Distribution}
\label{subsec:failure_distribution}
\begin{table}
    \centering
    \caption{The distribution of failure cases.}
    \label{tab:dist_failure}
    \resizebox{\linewidth}{!}{
    \begin{tabular}{l|cccc}
    \hline
    Datasets & Instruction Ambiguity & Perception Failure & Reasoning Failure \\
    \hline
    AirVLN-S & 45.4   & 34.1   & 20.5    \\
    % \hdashline
    AirVLN-E  & 5.8 & 76.8 & 17.4  \\
    \hline
    \end{tabular}
    }
\end{table}

As shown in Table~\ref{tab:dist_failure}, the failure distribution indicates that for AirVLN-S, the most failure cases come from the instruction ambiguity, while for AirVLN-E with fine-grained landmarks, the most failure cases come from the perception failure.

\end{document}